\documentclass[conference,11pt]{IEEEtran}
\IEEEoverridecommandlockouts

\usepackage{cite}
\usepackage{amsmath,amssymb,amsfonts}
\usepackage{algorithmic}
\usepackage{graphicx}
\usepackage{textcomp}
\usepackage{xcolor}
\usepackage{url}
\usepackage{array}
\usepackage{booktabs}
\usepackage{multirow}
\usepackage{placeins} 
\usepackage{tabularx}
\usepackage{array}
\usepackage{microtype} 
\usepackage{enumitem} 
\setlist[enumerate,itemize]{itemsep=0pt, topsep=3pt, partopsep=0pt}
\usepackage[
    letterpaper,
    top=0.76in,      
    bottom=1.01in,   
    left=0.64in,     
    right=0.64in,    
    columnsep=0.25in,
    includeheadfoot
]{geometry}
\def\BibTeX{{\rm B\kern-.05em{\sc i\kern-.025em b}\kern-.08em
    T\kern-.1667em\lower.7ex\hbox{E}\kern-.125emX}}
\begin{document}

\title{Edge-FIT: Federated Instruction Tuning of Quantized LLMs for Privacy-Preserving Smart Home Environments}

\author{
    \IEEEauthorblockN{Vinay Venkatesh}
    \IEEEauthorblockA{\textit{Senior IEEE Member} \\
    \textit{vinay.venkatesh@ieee.org}\\
    Mountain View, California, USA \\
    0009-0000-7824-4820}
\and
    \IEEEauthorblockN{Vamsidhar R Kamanuru}
    \IEEEauthorblockA{\textit{University of California, San Diego} \\
    \textit{vamsidharkamanuru@gmail.com}\\
    Santa Clara, USA\\
    0009-0009-5577-4517}
\\[1.5ex] 
    \IEEEauthorblockN{Nikita Kothari}
    \IEEEauthorblockA{\textit{Senior MTS} \\
    \textit{nikitakothari@ieee.org}\\
    California, USA\\
    0009-0008-3182-0550}
    
\and
    \IEEEauthorblockN{Lav Kumar}
    \IEEEauthorblockA{\textit{Manager, Software Engineering} \\
    \textit{lavgupta01@ieee.org}\\
    California, USA\\
    0009-0006-7742-4022}
}

\maketitle

\begin{abstract}
This paper proposes Edge-FIT (Federated Instruction Tuning on the Edge), a scalable framework for Federated Instruction Tuning (FIT) of Large Language Models (LLMs). Traditional Federated Learning (TFL) methods, like FedAvg, fail when confronted with the massive parameter size of LLMs \cite{b3}, \cite{b6}. Our Edge-FIT framework combines federated learning with 4-bit Quantized Low-Rank Adaptation (QLORA), mitigating the core issues of communication and computational overhead. We demonstrate this by filtering the general-purpose Databricks Dolly 15k dataset for the IoT domain. Experimental results show the Edge-FIT-tuned Llama 2 (7B) achieves an F1-Score of 0.89. We also demonstrate a viable trade-off using the 3.8B Phi-3-mini model, validating Edge-FIT as a scalable framework for decentralized LLM deployment on home compute gateways.
\end{abstract}

\begin{IEEEkeywords}
Federated Learning, Instruction Tuning, Edge-FIT, Llama 2, Phi-3, LoRA, QLoRA, IoT, Privacy, Edge Computing.
\end{IEEEkeywords}

\section{\textbf{Introduction}}
The proliferation of IoT devices in smart homes creates vast amounts of personal, domain-specific data. This includes high-risk data such as raw acoustic data from voice assistants, security sensor logs that infer user patterns, and unencrypted command histories. While Large Language Models (LLMs) possess the advanced reasoning and structured output generation (e.g., JSON) capabilities needed for true smart home automation \cite{b1}, \cite{b13}, \cite{b15}, training on this sensitive data presents a fundamental privacy challenge. Traditional centralized fine-tuning is incompatible with the IoT ecosystem, as it requires moving raw user data to a central server.

Traditional Federated Learning (TFL), defined by algorithms like FedAvg, emerged as the solution, allowing for collaborative training without sharing raw data \cite{b3}. However, TFL was designed for smaller models and fails when applied to modern LLMs due to two critical bottlenecks:

\begin{enumerate}
    \item \textbf{Prohibitive Communication Overhead:} TFL requires clients to transfer the entire model's weight updates, which can be gigabytes for models like Llama 2 (7B), overwhelming typical residential upload bandwidth \cite{b5}.
    \item \textbf{Prohibitive Computational Feasibility:} TFL assumes the client can train the full model, but fine-tuning a 7B parameter model requires massive VRAM (e.g., $>$70GB), which is impossible on edge devices \cite{b5}.
\end{enumerate}

To overcome these bottlenecks, Parameter-Efficient Fine-Tuning (PEFT) techniques have emerged. Low-Rank Adaptation (LoRA) is a prominent PEFT technique that freezes the pre-trained model weights and injects small, trainable "adapter" matrices \cite{b2}. QLORA further optimizes this by quantizing the frozen base model to 4-bits, making it possible to fine-tune massive models on consumer-grade GPUs \cite{b6}.

While recent work has focused on leveraging FL and LoRA for model personalization \cite{b18}, \cite{b19} and handling heterogeneous instruction sets \cite{b20} (as detailed in Section II), our work addresses a different, complementary challenge: \textbf{practical hardware feasibility} for building a \textbf{single, unified global model} in a specific domain. To this end, we propose \textbf{Edge-FIT} (Federated Instruction Tuning on the Edge).

Our specific contributions are:
\begin{itemize}[leftmargin=*]
    \item We propose a framework integrating FL with \textbf{QLoRA} to explicitly solve the dual bottlenecks of communication (LoRA) and client-side **computation (VRAM)**, distinct from PFL-focused approaches.
    \item We validate this framework using a novel \textbf{two-tiered hardware simulation} (high-VRAM A10 gateway vs. low-VRAM 8GB Jetson edge device) on a curated IoT dataset.
    \item We demonstrate that this approach achieves \textbf{near-centralized performance} (e.g., 0.80 F1 vs. 0.81 for Phi-3), offering a practical blueprint for deploying privacy-preserving LLMs on real-world edge hardware.
\end{itemize}

\section{\textbf{Related Work}}
Our work intersects with general federated learning, specifically on non-IID data \cite{b10}, and the emerging field of federated LLM tuning. Standard FL protocols \cite{b3} have been widely studied for IoT \cite{b8}, \cite{b7}, but traditional models lack the reasoning capabilities of LLMs.

The intersection of FL and PEFT is a highly active research area. Several frameworks have adopted the "FedLoRA" name to solve related, but distinct, challenges. For example, \textbf{pFedLoRA} \cite{b18} and the work by Wu et al. \cite{b19} focus on \textbf{Personalized Federated Learning (PFL)}. Their primary goal is to manage client heterogeneity by decomposing models into shared and personalized low-rank matrices. Similarly, the \textbf{FedIT} framework \cite{b20} (and its 'Shepherd' repository) uses FL and LoRA to build models from heterogeneous client instructions, also with a focus on personalization.

Another line of work, such as \textbf{FedQLoRA} \cite{b21}, addresses the algorithmic problem of \textbf{quantization bias} that arises when heterogeneous clients use different quantization levels. In contrast, our \textbf{Edge-FIT} framework is not focused on personalization or correcting quantization error. Rather, we address the \textbf{practical, systems-level challenge} of training a \textbf{single, robust, global model} using a standardized QLoRA approach \cite{b6} on resource-constrained hardware. Our architectural novelty is in using \textbf{QLoRA} \cite{b6} to solve the \textbf{dual bottlenecks} of communication (via LoRA) and the critical client-side \textit{computational (VRAM)} cost simultaneously. We validate this approach with a novel \textbf{two-tiered hardware simulation} (detailed in TABLE~\ref{tab:stacks}) to prove its feasibility in the Smart Home IoT domain.

\begin{figure*}[t]
    \centering
    \includegraphics[width=0.9\textwidth]{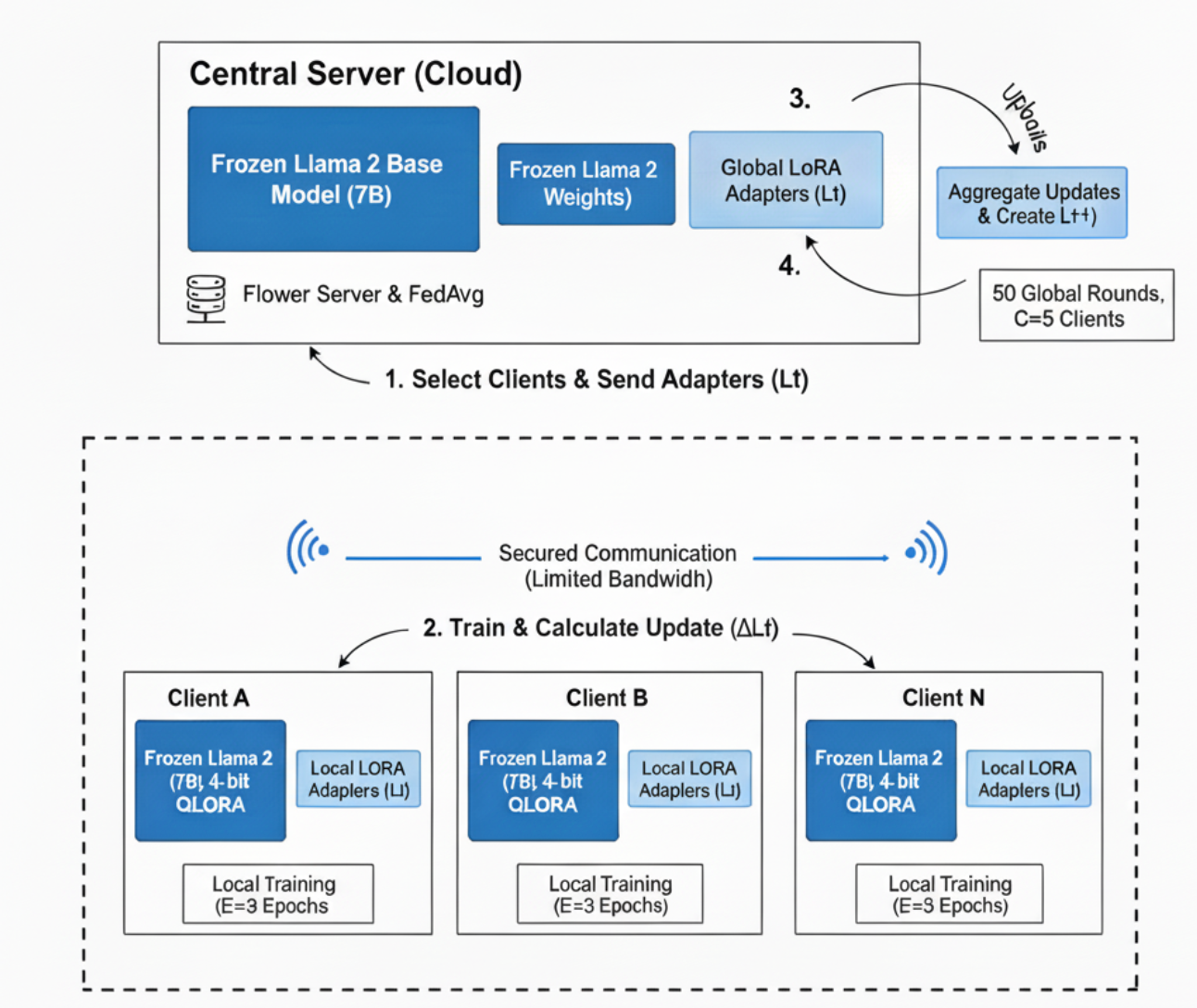} 
    \caption{Federated Instruction Tuning (FIT) using our proposed \textbf{Edge-FIT} framework. Clients only train and transmit the tiny adapter updates.}
    \label{fig:FIT}
\end{figure*}

\section{\textbf{Methodology}}
\subsection{\textbf{Model and Dataset Selection}}
We selected the Llama 2 (7B-chat) model as our primary LLM \cite{b1}. To analyze scalability, we also evaluated the Microsoft Phi-3-mini (3.8B) model. All models were sourced from their official Hugging Face repositories.

We used the Databricks Dolly 15k dataset as our instruction source \cite{b4}. The initial 15,011 pairs were filtered to create an IoT-specific dataset. We applied:

\begin{enumerate}
    \item \textbf{Exclusion Filters (Non-IoT):} Removed non-technical categories like "Creative Writing" and general "Closed QA" \cite{b4}.
    \item \textbf{Inclusion Filters (IoT Relevant):} Retained and augmented technical categories like "Information Extraction," "Classification," and "Brainstorming" with IoT-specific terms (e.g., "sensor data," "device log," "automation routine") \cite{b4}. An example of a filtered data point is shown in Table \ref{tab:my_example}.
\end{enumerate}

\subsection{\textbf{Example of Filtered and Formatted Data Point}}
The filtered records are formatted into the Llama 2 prompt template: [INST] \{instruction\} [/INST] \{response\}.

\begin{table}[h]
\caption{Example of Filtered IoT Data Point}
\label{tab:my_example}
\centering
\begin{tabularx}{\columnwidth}{|p{1.8cm}|X|} 
\hline
\textbf{Field} & \textbf{Example of Filtered IoT Data Point} \\
\hline
Category & Information Extraction (Augmented) \\
\hline
Instruction & \sloppy The user's motion sensor data shows activity in the living room at 3:00 AM. Classify this event as 'Normal' or 'Anomaly' and explain why. \\
\hline
Response & \sloppy Anomaly. The event occurred outside standard waking hours, suggesting an unusual security or device malfunction event. \\
\hline
\end{tabularx}
\end{table}

\subsection{\textbf{Data Generation and Pre-processing}}
The initial filtering yielded $\sim$4,500 base samples. We then used the Self-Instruct method \cite{b14} with Llama 2 (7B) as the "teacher" to generate an additional 2,500 new IoT-specific instruction-response pairs \cite{b1}. This process, which used a few-shot prompt (e.g., classifying sensor logs and defining JSON routines), resulted in a final dataset of $\sim$7,000 high-quality samples.

This final dataset was partitioned among $N=10$ simulated clients. To mimic real-world data heterogeneity, we used a Dirichlet distribution $Dir(\alpha)$ with $\alpha=0.3$ to create a non-IID "Label Skew," where each client has a statistically unique distribution of instruction types \cite{b10}.

\subsection{\textbf{Federated Training Strategy}}
Our strategy combines 4-bit QLoRA \cite{b6} (with $r=16$) with FedAvg \cite{b3}. We run $T=50$ global communication rounds, with $C=5$ clients sampled per round. Clients upload only the lightweight LoRA adapter updates ($\Delta \mathbf{L}_i$), which the server aggregates:

\begin{flalign}
& \mathbf{L}^t \leftarrow \mathbf{L}^{t-1} + \sum_{i=1}^{C} \frac{|\mathcal{D}_i|}{\sum_{j=1}^{C} |\mathcal{D}_j|} \cdot \Delta \mathbf{L}_i &
\end{flalign}

\subsection{\textbf{Federated Training Loop and Implementation}}
This simulation executes the TFL protocol using the Flower framework \cite{b12}. The overall architecture, shown in Fig. \ref{fig:FIT}, consists of 50 iterations of the following loop:

\begin{enumerate}
    \item \textbf{Server Initialization \& Client Spawning:} The server process initiates a Flower server, loads the FedAvg strategy, and initializes global adapters $\mathbf{L}^0$. It also launches 10 parallel client processes, mapping each to its non-IID data partition.
    \item \textbf{Selection \& Distribution:} At the start of a round $t$, the server selects 5 clients ($C=5$) and sends them the identical global adapters $\mathbf{L}^t$.
    \item \textbf{Client-Side Training:} Each client loads the adapter $\mathbf{L}^t$ onto its frozen, 4-bit quantized base model and trains it for $E=3$ local epochs on its private data. This modifies only the LoRA adapter weights, resulting in $\mathbf{L}^{t+1}\_A$.
    \item \textbf{Upload:} After training, the client calculates the difference $\Delta \mathbf{L}\_A = \mathbf{L}^{t+1}\_A - \mathbf{L}^{t}\_A$. The client returns only this tiny update to the server.
    \item \textbf{Aggregation:} The Flower server waits for all 5 client updates and applies Equation 1 to calculate the weighted average, resulting in the new global adapter, $\mathbf{L}^{t+1}$.
\end{enumerate}

\section{\textbf{SYSTEM ARCHITECTURE AND BASELINES}}
The simulation was deployed on AWS, with the hardware and software stack detailed in TABLE~\ref{tab:stacks}. 

\subsection{\textbf{PEFT and FL Configuration}}
To make the models feasible, we used 4-bit QLoRA \cite{b6}. The base models were loaded in a compressed 4-bit data type (NF4), and trainable LoRA adapters ($r=16$) were injected into the attention layers \cite{b2}. All local training used a standard \textbf{causal language modeling loss function}. For both the Llama 2 and Phi-3 experiments, all hyperparameters were kept identical:
\begin{itemize}[leftmargin=*]
    \item Global Rounds (T): 50
    \item Clients Sampled (C): 5
    \item Local Epochs (E): 3
    \item LoRA Rank (r): 16
    \item Optimizer: AdamW
    \item Learning Rate: 2e-4
\end{itemize}

\begin{table*}[t]
    \centering
    \caption{HARDWARE AND SOFTWARE STACK}
    \label{tab:stacks}
    \begin{tabular}{|p{2.5cm}|p{4.5cm}|p{3.5cm}|p{5.5cm}|} 
    \hline
    \textbf{Component Category} & \textbf{Specification} & \textbf{Access} & \textbf{Role in the experiment} \\
    \hline
    \textbf{Hardware (Server)} & AWS EC2 p3.8xlarge Instance (4x NVIDIA V100 16GB) & Rented from AWS & \textbf{Central Aggregator:} Hosts Flower server, orchestrates FL rounds, and performs FedAvg. \\
    \hline
    \textbf{Hardware (Client)} & NVIDIA A10 (24GB) & Simulated via AWS EC2 G5 & \textbf{Local Client (Gateway)}: The $\ge$24GB VRAM positions this as a "home server" or "compute gateway." \\
    \hline
    \textbf{Hardware (Client)} & NVIDIA Jetson Orin Nano (8GB) & Simulated via Greengrass & \textbf{Local Client (Edge)}: Simulated to run Phi-3-mini, mimicking the 8GB memory of an edge device. \\
    \hline
    \textbf{Base Models} & Llama 2 (7B-chat), Phi-3-mini (3.8B) & Hugging Face & \textbf{Frozen Knowledge Base}: Hosted on clients for local training. \\
    \hline
    \textbf{ML Framework} & PyTorch 2.1.0 (with CUDA 12.1) & Open Source & Executes all local training and inference logic. \\
    \hline
    \textbf{FL Framework} & Flower (flwr) 1.8.0 & Open Source & \textbf{Orchestration}: Manages the simulation, client sampling, and aggregation protocol [12]. \\
    \hline
    \textbf{PEFT Libraries} & peft 0.5.3, bitsandbytes 0.41.3, transformers 4.35.2 & Open Source & \textit{peft} injects LoRA adapters, \textit{bitsandbytes} enables 4-bit QLoRA quantization. \\
    \hline
    \textbf{Dataset} & Databricks Dolly 15k & Open Source [4] & The raw data source for filtering and augmentation. \\
    \hline
    \end{tabular}
\end{table*}

\subsection{\textbf{Experimental Baselines}}
Our experimental baselines validate the \textbf{architectural feasibility} of the Edge-FIT framework. The goal is to show Edge-FIT closes the gap to the ideal centralized model while providing privacy, thus validating the framework's primary objective.

We evaluated four distinct procedures. Evaluation metrics (Accuracy, Precision, Recall, F1-Score) are calculated for the classification tasks within our held-out test set.

\begin{enumerate}
    \item \textbf{Base LLM (Zero-Shot)}: The out-of-the-box model performance.
        \begin{itemize}[leftmargin=*]
            \item \textbf{Llama 2}: The 4-bit quantized model was loaded on an NVIDIA A10 (24GB) for inference.
            \item \textbf{Phi-3}: The 4-bit quantized model was loaded on a simulated 8GB client for inference.
        \end{itemize}
        
    \item \textbf{Local-Only (Avg.)}: Simulates clients training only on their own siloed, non-IID data.
    \begin{itemize}[leftmargin=*]
            \item Ten independent QLoRA fine-tunes were run on the 10 respective clients for each model.
            \item The reported metric is the average F1-score of these 10 models evaluated on the full test set.
        \end{itemize}
    
    \item \textbf{Centralized Fine-Tuning (Ideal Baseline)}: The "gold standard" non-private baseline, where a single QLoRA fine-tuning was run on the entire aggregated dataset.
    \begin{itemize}[leftmargin=*]
            \item Training was run on a single NVIDIA V100 GPU for both Llama 2 and Phi-3.
        \end{itemize}
    
    \item \textbf{Edge-FIT (Proposed)}: Our full federated framework, run for T=50 rounds.
    \begin{itemize}[leftmargin=*]
            \item Llama 2 clients trained on A10s; Phi-3 clients on simulated 8GB devices.
            \item The final aggregated global adapter ($\mathbf{L}^{50}$) was evaluated on the held-out test set.
        \end{itemize}
\end{enumerate}

\section{\textbf{RESULTS AND DISCUSSION}}
The comparative performance of all Llama 2 (7B) baselines is presented in TABLE~\ref{tab:metrics}.

\begin{table*}[t]
    \centering 
    \caption{Comparative Performance Metrics (LLAMA 2 7B)}
    \label{tab:metrics}
    \begin{tabularx}{\textwidth}{|l|X|c|c|c|c|}
        \hline
        \textbf{Model Variant} & \textbf{Training Method} & \textbf{Accuracy} & \textbf{Precision} & \textbf{Recall} & \textbf{F1-Score}  \\
        \hline
        Base Llama 2 (7B) & Zero Shot/ No Training & 0.31 & 0.28 & 0.35 & 0.31 \\
        \hline
        Local-Only (Avg.) & Individual Training & 0.78 & 0.65 & 0.88 & 0.75 \\
        \hline
        Centralized Fine-Tuning & Full Data (Ideal Baseline) & 0.93 & 0.94 & 0.92 & 0.93 \\
        \hline
        \textbf{Edge-FIT (Proposed)} & \textbf{FedAvg on LoRA Adapters} & \textbf{0.89} & \textbf{0.87} & \textbf{0.91} & \textbf{0.89} \\
        \hline
    \end{tabularx}
\end{table*}

\subsection{\textbf{Model-Size and VRAM Performance Trade-off}}
To provide a pathway for edge deployment, we replicated the experiments using the smaller Phi-3-mini (3.8B) model. This analysis compares the VRAM requirements and achievable performance. These results are shown in TABLE~\ref{tab:phi}.

\begin{table*}[t]
    \centering
    \caption{MODEL-SIZE AND VRAM PERFORMANCE TRADE-OFF}
    \label{tab:phi}
    \begin{tabularx}{\textwidth}{|l|X|c|c|c|}
        \hline
        \textbf{Model Variant} & \textbf{Client VRAM Req. (4-bit)} & \textbf{Base F1 (Zero-Shot)} & \textbf{Centralized F1} & \textbf{Edge-FIT (FedAvg) F1} \\
        \hline
        Llama 2 (7B) & $\geq$24 GB (System) & 0.31 & 0.93 & 0.89\\
        \hline
        Phi-3-mini (3.8B) & 8 GB (System)  & 0.22 & 0.81 & 0.80\\
        \hline
    \end{tabularx}
\end{table*}

\subsection{\textbf{Discussion}}
The results from our experiments validate the Edge-FIT framework and lead to four key insights:

\begin{enumerate}
    \item \textbf{The VRAM Bottleneck is Solved for Practical Models:} The Phi-3-mini result is highly significant. It proves that by using QLoRA, it is possible to federatively fine-tune a powerful, multi-billion parameter LLM on devices with only 8GB of VRAM. This overcomes the primary hardware obstacle preventing LLM deployment on real-world edge devices.
    \item \textbf{Collaboration Overcomes Data Silos:} The 14-point F1-score improvement of Edge-FIT (0.89) over the Local-Only average (0.75) for Llama 2 demonstrates the power of federation. This jump occurs because FedAvg allows clients to benefit from patterns in other clients' non-IID data, overcoming the "local myopia" of isolated training.
    \item \textbf{Quantifying the Privacy-Utility Trade-off:} The proximity of the Edge-FIT model's F1-Score (0.89) to the Centralized Baseline (0.93) is a key finding. It serves as empirical evidence that near-optimal utility can be achieved with a minimal performance trade-off, all while maintaining the strict privacy guarantee of keeping raw user data localized.
    \item \textbf{A Blueprint for Scalable Edge AI:} The combined results of the two-tiered simulation provide a practical blueprint for real-world deployment. Organizations can use powerful models like Llama 2 for less-constrained "gateway" devices, while deploying capable models like Phi-3-mini on low-power end devices.
\end{enumerate}

\section{\textbf{Advantages and Tradeoffs of the Edge-FIT Method}}
\subsection{\textbf{Advantages over Traditional Federated Learning (TFL)}}
\begin{enumerate}
    \item \textbf{Feasibility Solution:} Our framework solves the TFL communication bottleneck. Edge-FIT uses PEFT to reduce the communication size by $>99.9$\% \cite{b6}.
    \item \textbf{Capability Shift:} LLMs unlock complex, unstructured language understanding and structured command generation (e.g., JSON) for autonomous IoT systems \cite{b13}, \cite{b15}.
    \item \textbf{Privacy by Design:} Raw, sensitive user data remains on the local device, mitigating privacy risks \cite{b11}, \cite{b7}.
    \item \textbf{Reduced Bandwidth:} Only the lightweight LoRA adapters are transmitted \cite{b6}, \cite{b5}.
    \item \textbf{Scalability:} The system distributes the computational burden across multiple edge devices \cite{b7}.
\end{enumerate}

\subsection{\textbf{Trade-offs (Limitations)}}
\begin{enumerate}
    \item \textbf{Non-IID Data Convergence:} The primary challenge is that clients hold Non-IID data \cite{b10}. This can introduce 'model drift.' While our FedAvg approach shows high performance, convergence can be slower than on a centralized dataset.
    \item \textbf{Increased Complexity:} Implementing FL requires a specialized coordination framework (like Flower) \cite{b12} and managing heterogeneous client states.
    \item \textbf{On-Device Computational Burden:} Although PEFT dramatically reduces resource usage, local training still requires significant VRAM (i.e., the need for $\ge$24 GB VRAM in the simulation).
\end{enumerate}

\section{\textbf{Conclusion and Future Work}}
This paper proposed \textbf{Edge-FIT}, a framework for Federated Instruction Tuning that makes training LLMs in privacy-sensitive IoT environments feasible. By combining 4-bit QLoRA with federated averaging, our approach proved both communication- and computation-efficient. Experimental results showed our Edge-FIT-tuned Llama 2 (7B) model (0.89 F1) achieves performance remarkably close to the non-private centralized baseline (0.93 F1). We also demonstrated that smaller models like Phi-3-mini (0.80 F1) provide a practical pathway for deployment on lower-VRAM edge hardware. This work validates Edge-FIT as a feasible, two-tiered strategy for deploying advanced, privacy-preserving AI in smart homes.

However, this work is a simulation with key limitations that define our future work:

\begin{enumerate}
    \item \textbf{Simulation vs. Reality:} Our synchronous simulation did not model real-world factors like client drop-out, network latency, or hardware heterogeneity.
    \item \textbf{Robust Aggregation:} We used FedAvg as a baseline. A critical next step is to test robust aggregation algorithms (e.g., FedProx) to mitigate model drift from non-IID data \cite{b10}.
    \item \textbf{Direct SOTA Comparison:} Future work should include a direct empirical benchmark of Edge-FIT against recent personalization-focused frameworks \cite{b18, b19, b20} on a common task.
    \item \textbf{Enhanced Security:} Future work should integrate techniques like Differential Privacy (DP) \cite{b16} or Secure Aggregation.
    \item \textbf{System Architecture:} We will also explore Hierarchical FL architectures to further reduce latency \cite{b17}.
\end{enumerate}

\end{document}